\documentclass[conference]{article}
\usepackage[margin=1in]{geometry}

\usepackage[utf8]{inputenc}
\usepackage[T1]{fontenc}
\usepackage[english]{babel}
\usepackage{graphicx}
\usepackage{amssymb}
\usepackage{amsthm}
\usepackage{amsmath}
\usepackage{amsfonts}
\usepackage{color}

\newcommand{\mc}{\mathcal}
\newcommand{\R}{\mathbb{R}}
\newcommand{\norm}[2]{\left\Vert #1 \right\Vert_{{#2}}}
\DeclareMathOperator*{\argmin}{arg\,min}
\newcommand{\prob}{\mathbb{P}}
\newcommand{\EE}{\mathbb{E}}
\newcommand{\smean}{\overline}

\newcommand{\frmu}{\mu}
\newcommand{\graphdist}{{\rm d}}
\newcommand{\gramiam}{C}
\newcommand{\dzvector}[1]{\mathbf{\boldsymbol{#1}}}
\renewcommand{\vec}[1]{\dzvector{#1}}

\DeclareMathOperator{\arl}{ARL}

\usepackage{authblk}
\title{Anomaly and Change Detection in Graph Streams through Constant-Curvature Manifold Embeddings}
\author[1]{Daniele Zambon}
\author[2]{Lorenzo Livi}
\author[1,3]{Cesare Alippi}
\affil[1]{\small Faculty of Informatics, Universit\`{a} della Svizzera italiana, Switzerland, Email: \texttt{daniele.zambon@usi.ch}, \texttt{cesare.alippi@usi.ch}.}
\affil[2]{\small Department of Computer Science, College of Engineering, Mathematics and Physical Sciences, University of Exeter, United Kingdom, Email: \texttt{l.livi@exeter.ac.uk}.}
\affil[3]{\small Department of Electronics, Information, and Bioengineering, Politecnico di Milano, Italy, Email: \texttt{cesare.alippi@polimi.it}.}

\date{April 27, 2018}

\newcommand\blfootnote[1]{%
  \begingroup
  \renewcommand\thefootnote{}\footnote{#1}%
  \addtocounter{footnote}{-1}%
  \endgroup
}

\def\copyrightnotice{%
\copyright 2018 IEEE. Personal use of this material is permitted. Permission
from IEEE must be obtained for all other uses, in any current or future
media, including reprinting/republishing this material for advertising or
promotional purposes, creating new collective works, for resale or
redistribution to servers or lists, or reuse of any copyrighted
component of this work in other works.\hfill}

\begin{document}

\maketitle
\blfootnote{\copyrightnotice}


    


\begin{abstract}
Mapping complex input data into suitable lower dimensional manifolds is a common procedure in machine learning.
This step is beneficial mainly for two reasons: (1) it reduces the data dimensionality and (2) it provides a new data representation possibly characterised by convenient geometric properties.
Euclidean spaces are by far the most widely used embedding spaces, thanks to their well-understood structure and large availability of consolidated inference methods. 
However, recent research demonstrated that many types of complex data (e.g., those represented as graphs) are actually better described by non-Euclidean geometries.
Here, we investigate how embedding graphs on constant-curvature manifolds (hyper-spherical and hyperbolic manifolds) impacts on the ability to detect changes in sequences of attributed  graphs.
The proposed methodology consists in embedding graphs into a geometric space and perform change detection there by means of conventional methods for numerical streams.
The curvature of the space is  a parameter that we learn to reproduce the geometry of the original application-dependent graph space.
Preliminary experimental results show the potential capability of representing graphs by means of curved manifold, in particular for change and anomaly detection problems.
\end{abstract}

\section{Introduction}
\label{sec:intro}

Not all data structures are well-described by Euclidean geometry. For instance, important examples include hyperbolic geometry underlying complex networks \cite{wu2015emergent,boguna2010sustaining}, numeric features augmented with pairwise relations \cite{poincare_embedding}, and irregular/curved shapes typical of computer vision \cite{bronstein2016geometric,straub2015dirichlet}.

Here, we elaborate over our previous contribution on change detection in sequences of attributed graphs \cite{2017arXiv170606941Z}. We propose and experimentally evaluate the effectiveness of using non-Euclidean embedding spaces for graphs in anomaly and change detection problems. 
More precisely, in addition to the embedding based on dissimilarity representation, e.g, see \cite{2017arXiv170606941Z}, here we take into account mappings into three additional embedding spaces characterised by a constant curvature: (i) Euclidean $d$-space, (ii) $d$-sphere and (iii) $d$-dimensional hyperbolic space \cite{bridson2013metric}. 
Constant-curvature manifolds allow the designer to easily learn the curvature $\kappa\in \R$ from the application data, which is controlled by a single scalar parameter. 
In turn, this allows to identify the appropriate embedding as (i) Euclidean, $\kappa = 0$; (ii) spherical, $\kappa > 0$; and (iii) hyperbolic, $\kappa < 0$.
We denote a generic manifold with constant curvature $\kappa$ as $\mathbb M_\kappa$.
In particular, we show that the family of graphs considered in this paper is better embedded in a hyperbolic manifold than in a spherical one.

In this paper, we address the problem of monitoring a stream $g_1,g_2,\dots,g_t,\dots$ of i.i.d.\ attributed graphs generated by process $\mc P$ according to nominal distribution $Q$. 
We say that a change in stationarity occurs at change time $\tau$ if graph $g_t$ for $t\geq \tau$ is drawn according to non-nominal distribution $\widetilde Q\neq Q$, i.e.,
$$
\mc P\ :\ 
\begin{cases}
g_t \sim Q            & t <  \tau\\
g_t \sim \widetilde Q & t\geq\tau.
\end{cases}
$$
In the same framework, an anomalous event is a graph $g_\tau$ which is very unlikely to have been generated by distribution $Q$, given a confidence level. 
The proposed monitoring methodology for anomaly and change detection is the same, the difference is the final statistic adopted for the detection (see Section \ref{sec:detection-test}).

In the sequel, we consider the space of attributed graphs $\mc G$ and a distance $\graphdist(\cdot,\cdot)$ between two graphs (e.g., the graph edit distance \cite{riesen2013novel}). Since, the space of attributed graphs $\mc G$ is rarely described by an Euclidean geometry, acting directly on $\mc G$ is non-trivial and usually computationally expensive due to the cost associated with the 
graph matching procedures \cite{gm_survey,emmert2016fifty}.
In order to process graphs for change detection, this paper proposes an embedding procedure where the generic graph $g_t\in\mc G$ is mapped onto a point lying on a constant-curvature manifold $\mathbb M_\kappa$. A numeric stream $x_1,x_2,\dots,x_t,\dots$ is hence generated by the embedding, through map $f(\cdot):g_t \mapsto x_t$, and a detection test is finally applied to the $x$-stream.

The present paper contributes in exploiting embedding of graphs onto Riemannian manifolds in order to, subsequently, apply anomaly and change detection tests on the manifold; it proposes also a prototype-based embedding map which can also deal with the out-of-sample problem, that is, the problem of embedding newly observed graphs without system reconfiguration. 

The paper is structured as follows.
Section \ref{sec:manifolds} introduces the considered constant-curvature manifolds.
In Section \ref{sec:embedding}, we describe the embedding techniques and how they are extended in the out-of-sample case.
Section \ref{sec:detection-test} provides details on the anomaly and change detection tests taken into account in this paper. Preliminary experiments and related discussions follow in Section \ref{sec:experiments}. 
Conclusions and future directions are provided in Section \ref{sec:conclusions}.

\section{Constant-curvature manifolds}
\label{sec:manifolds}

A Riemannian manifold is a differential manifold with a scalar product defined on the tangent bundle. By means of the scalar product, a metric distance is derived. We denote a generic manifold with $\mathbb M = (\mc M, \rho(\cdot,\cdot) )$, where $\mc M$ is the domain space and $\rho(\cdot,\cdot):\mc M\times \mc M \rightarrow \R_+$ is the geodesic distance between two points in $\mc M$ \cite{bridson2013metric}. 
Manifold $\mathbb M$ is also locally homeomorphic to a $d$-dimensional Euclidean space; accordingly, we say that the dimension of the manifold is $d$. 
$\mathbb M_\kappa = (\mc M_\kappa, \rho_\kappa(\cdot,\cdot) )$ is a special family of Riemannian manifolds characterised by a constant curvature $\kappa$ that defines the shape of the manifold.

We explore three constant-curvature manifolds by acting on $\kappa$: Euclidean, $\kappa=0$; spherical, $\kappa>0$; and hyperbolic, $\kappa<0$. Manifold $\mathbb M_{\kappa=0}$ is the usual $d$-dimensional Euclidean space equipped with the metric distance
\begin{equation}
\label{eq:euclidean-distance}
    \rho_{\kappa}(x,y)=\norm{x-y}{2} = \sqrt{\langle x,x\rangle + \langle y,y\rangle - 2 \langle x,y\rangle},
\end{equation}
where $\langle x,y\rangle$ is the inner product $\langle x,y\rangle = \sum_i x(i)y(i)$.

The $d$-dimensional spherical manifold $\mathbb M_\kappa$ ($\kappa>0$) has support on $d$-sphere of radius $r=\frac{1}{\sqrt{\kappa}}$
\begin{equation*}
    \mc M_\kappa:=\left\{ x\in\R^{d+1}\ |\ \langle x,x \rangle = r^2\;\right\}.
\end{equation*} 
As a distance between two points $x,y\in\mc M_\kappa$, the geodesic metric distance is
\begin{equation}
\label{eq:spherical-distance}
    \rho_{\kappa}(x,y):= r \cos^{-1}\left(\frac{\langle x,y\rangle}{r^2}\right).
\end{equation}

Hyperbolic manifolds are constructed over a $d+1$-dimensional pseudo-Euclidean space equipped with the scalar product%
\footnote{Being $\langle \cdot,\cdot \rangle$ not positive definite, this is not a typical inner product.
With little abuse of notation, we denote by $\langle\cdot,\cdot\rangle$ both the Euclidean and pseudo-Euclidean scalar products. We believe this fact will not mislead the reader, since it should be clear by the curvature which scalar product is involved.}
\begin{equation*}
\langle x,y\rangle := \sum_{i=2}^{d+1} x(i)^2-x(1)^2.
\end{equation*}
Similarly to the spherical case, for every $\kappa<0$, we define the hyperbolic manifold as
$$\mc M_\kappa:=\left\{ x\in\R^{d+1}\ |\ \langle x,x \rangle = - r^2,\;x(1)>0 \right\},$$
with $r=\frac{1}{\sqrt{-k}}$. The geodesic metric distance is
\begin{equation}
\label{eq:hyperbolic-distance}
    \rho_{\kappa}(x,y):= r \cosh^{-1}\left(\frac{-\langle x,y\rangle}{r^2}\right).
\end{equation}
Quantity $r$ plays an analogous role of the radius in the spherical case; as such, we will call it \emph{radius} also in the hyperbolic case.

As for conventional Euclidean spaces, the distance between two points on constant-curvature manifolds can be obtained by means of the scalar product. Further, we can represent the scalar product $\langle x,y\rangle$ as a matrix multiplication in the form $x^\top I_\kappa y$, where  $I_\kappa$ is the identity matrix for $\kappa\geq 0$, and the identity matrix except for element in position $(1,1)$ set to $-1$ in the $k<0$ case.

\section{Embedding based on dissimilarity matrix}
\label{sec:embedding}

The proposed methodology consists in mapping a sequence of attributed graphs $g_1,g_2,\dots,g_t,\dots$, $g_t\in\mc G$, into a sequence of numeric vectors, $x_1,x_2,\dots,x_t,\dots$, $x_t\in\mc M_k$.
By following \cite{wilson2014spherical}, we consider a symmetric distance measure $\graphdist(\cdot,\cdot)$ operating in $\mc G$ and propose to adopt an embedding aiming at preserving the distances, i.e.,
\begin{equation}
\label{eq:similar-distances}
\rho_\kappa(f(g_1),f(g_2)) \approx \graphdist(g_1,g_2) \qquad \forall g_1,g_2\in\mc G.
\end{equation}
$f(\cdot)$ is the embedding map $\mc G \rightarrow \mc M_\kappa$ and $\rho_\kappa(\cdot, \cdot)$ is the corresponding distance operating in $\mc M_\kappa$; see Equations 
\eqref{eq:euclidean-distance}, \eqref{eq:spherical-distance} and \eqref{eq:hyperbolic-distance}.

The method configuration phase is composed of three, conceptually different, steps.
In the first one, we learn the underlying manifold by processing graphs taken from a finite training set $T$.
Since there is a one-to-one correspondence between the curvature $\kappa\in\R$ and manifolds introduced above, this step aims at estimating curvature $\kappa$ for the application at hand.
In the second step, we select a set of embedded points associated with $T$, say $X$, satisfying \eqref{eq:similar-distances}.
In other terms, the distances between pairs of vectors should match those of the original graphs as much as possible (see \eqref{eq:similar-distances}).
The third step addresses the out-of-sample problem, that is, how to apply map $f(\cdot)$ to all graphs $g\in\mc G$ not in the training set. In order to solve this problem, we propose here to adopt a prototype-based technique, where prototypes serve as landmarks to find the corresponding position of graphs on $\mc M_\kappa$.
Next subsections will discuss the three steps.

\subsection{Embedding the training set}
\label{sec:training_set_embedding}

The first two steps of the procedure, namely defining a curvature $\kappa$ and determining a configuration $X=\{x_n\}_1^N\subseteq \mc M_\kappa$ for training graphs $T=\{g_n\}_1^N$, are carried out simultaneously; as such, they are treated as a single step.

Let us arrange the vectors of configuration $X$ row-wise 
so that set $X$ is represented as the matrix $X=[x_1|\dots|x_N]^\top$. Let $D=\graphdist(T,T)$ be the square matrix containing the pair-wise distances between graphs in $T$, in the sense that generic element $D(i,j)$ of $D$ is $\graphdist(g_i,g_j)$.

As mentioned in Section~\ref{sec:manifolds}, the distance $\rho_\kappa(x_i,x_j)$ can be derived by the scalar product $\langle x_i,x_j \rangle$, as such we consider matrix $\gramiam$, collecting all pair-wise scalar products.
Matrix $\gramiam$ can be written as $\gramiam=X I_\kappa X^\top$; however, due to the relation between distances and scalar products, matrix $\gramiam$ can be computed (Equations \eqref{eq:euclidean-distance}, \eqref{eq:spherical-distance}, \eqref{eq:hyperbolic-distance}) as
\begin{equation}
\label{eq:corr-matrix}
    \gramiam=
    \begin{cases}
        r^2\cdot \cos(D/r) & \mathbb M_{\kappa>0},\\
        -\frac{1}{2}(J\cdot D^{\circ 2} \cdot J) & \mathbb M_{\kappa=0},\\
        -r^2\cdot \cosh(D/r) & \mathbb M_{\kappa<0},
    \end{cases}
\end{equation}
where $\circ p$ represents the exponentiation operator to power $p$ and operations $\cos(A)$, $\cosh(A)$, and $A^{\circ p}$ are applied component-wise to matrix $A$. $J$ is the centering matrix. 
Further details can be found in \cite{pekalska2005dissimilarity,wilson2014spherical}.

Ideally, we would like to solve equation $\gramiam = X I_\kappa X^\top$ w.r.t.\ $X$ to select embedded points $X$, where $\gramiam$ is given by \eqref{eq:corr-matrix}. In a practical applications, however, obtaining isometric embeddings is difficult and requires instead to solve the optimisation problem 
\begin{equation}
\label{eq:common-opt-problem}
    \argmin_X \norm{ X I_\kappa X^\top -\gramiam }{F}
\end{equation}
subject to the constraint that every $x \in X$ lays on $\mc M_k$.
\eqref{eq:common-opt-problem} can be solved thanks the eigen-decomposition $U \Lambda U^\top$ of symmetric matrix $\gramiam$. In the sequel, we adopt the convention that computed diagonal matrix $\Lambda$ stores the eigenvalues in ascending order. 
We need to solve \eqref{eq:common-opt-problem} by assuming a null curvature space at first and, then, the not null curvature case.

In case of null curvature, the problem can be solved by means of the classical Multi-Dimensional Scaling (MDS) \cite{pekalska2005dissimilarity}.
The vector configuration of the solution in hence
$$
X=U_d \Lambda_d^{\circ \frac{1}{2}},
$$
where $U_d$ and $\Lambda_d$ are the matrices reduced to the $d$-largest eigenvalues%
\footnote{Negative eigenvalues can be discarded. It should be commented that their presence is a strong indicator that data do not well fit with an Euclidean space.}.
Here the problem is unconstrained and $X$ can be any matrix in $\R^{N\times d}$.

To solve \eqref{eq:common-opt-problem} in curved spaces, we exploit a different formulation 
\begin{equation*}
    \argmin_X \norm{ B -\Lambda }{F}
\end{equation*}
which is an equivalent formulation of \eqref{eq:common-opt-problem} where $B=U^\top X I_\kappa X^\top U$.
Assuming $B$ to be diagonal, with diagonal vector $b$, the problem becomes
\begin{equation}
\label{eq:opt-problem-b}
    \argmin_b \norm{b-\lambda}{2},
\end{equation}
where $\lambda$ is the diagonal vector containing the eigenvalues of $\Lambda$.
Solution of \eqref{eq:opt-problem-b} depends on the sign of curvature $\kappa$, which also requires the application of different constraints on $b$ \cite{wilson2014spherical}.

In particular, for $\kappa>0$, it is required that $b(i)\geq 0$ holds for all $i$ and $U^{\circ 2} b = \frac{1}{k}\vec 1$.  The solution vector lays on the line passing through $\lambda$ and $\frac{1}{k}\vec 1$, and can be always provided in closed form.
Finally, a solution to \eqref{eq:common-opt-problem}, given \eqref{eq:opt-problem-b}, is obtained by dropping the smallest components of $b$ and reconstructing $X$ from it, i.e., $X=U_{d+1}B_{d+1}^{\circ \frac{1}{2}}$. 

In the hyperbolic case, \eqref{eq:opt-problem-b} can be formulated likewise with the difference that we need to take into account constraints coming from the pseudo-Euclidean geometry. Optimization problem \eqref{eq:opt-problem-b} is formulated in the same way.  
The constraint is $b(i)\geq 0$ for all $i>1$, $b(1)\leq 0$, and $U^{\circ 2} b = \frac{1}{k}\vec 1$. 
Finding a solution to the original problem \eqref{eq:common-opt-problem} by means of \eqref{eq:opt-problem-b} is not always possible as in the spherical case; this is the case when $\lambda_2\leq\frac{1}{k}$, where the line passing through $\lambda$ and $\frac{1}{k}\vec 1$ does not intersect the feasible set. 
Except for that case, a solution is always found in closed form on the line touching $\lambda$ and $\frac{1}{k}\vec 1$. 
A solution $X$ is constructed from $b$ and $U$ by dropping the smallest components of $b$ except for $b(1)$; therefore, the solution is $X=U_{d+1}|B_{d+1}|^{\circ \frac{1}{2}}I_\kappa$. 

In both spherical and hyperbolic cases, dropping some components of $b$ might lead to a vector configuration $X$ lying outside the feasible set. When data-point are close to set $\mc M$, this can be mitigated by projecting the points back on the manifold. 

The procedure described in this section is performed for a fixed radius $r$. 
A possible way to identify a suitable curvature $\kappa$ is to compare the resulting distance matrix $D_{\kappa}=\rho_\kappa(X,X)$ with matrix $D=\graphdist(T,T)$, then, a suitable $\kappa$ is selected as the one minimising the quantity
\begin{equation}
\label{eq:embedding-distortion}
    \hat \kappa = \argmin_{\kappa\in\R} \norm{D_{\kappa} -D}{F}.
\end{equation}
Such a measure estimates the distortion of training set $T$ and guides the designer towards the most appropriate embedding.

\subsection{Out-of-sample embedding}
\label{sec:test_data_embedding}

During the test (operational) phase of the method, every time a new graph $g_t$ is observed, we propose to map it onto a vector $x_t$ lying on the manifold $\mc M$.
Here, we consider a set of suitably chosen landmarks (prototypes) on the manifold $\mc M$, and find the position $x_t$ of the new graph by means of them. 
The prototype set $\{r_m\}_{m=1}^M=R$ is composed by $M$ graphs $r_m\in\mc G$. Set $R$ is associated with a set $X_R$ collecting the corresponding positions on the manifold; As done before, we assume $X_R$ to be in matrix form with each row $x_m\in\mc M$ being given by $f(r_m)$.

Let us assume set $R\subseteq T$ is given. In order to embed graph $g_t$, the first step consists in computing its dissimilarity representation $y_t = \graphdist(g_t,R)$: the generic $m$-th component of $y_t$ is the graph distance $y_t(m)=\graphdist(g_t,r_m)$.
Once $y_t$ is computed, the embedded vector is
$$ 
x_t = \argmin_{x\in\mc M} \norm{X_R I_\kappa x-\gramiam_R}{2}^{2},
$$
where $\gramiam_R$ is the vector obtained by applying \eqref{eq:corr-matrix} to $y_t$. 

The selection of the prototype set $R$ is performed only once during the training phase and, hence, does not change during the operational one.
We adopt the $k$-centres algorithm on the configuration $X$. The algorithm covers the configuration points in $X$ using balls of equal radius centred in the prototype images $f(r_m)$; this is done by minimising the maximal distance -- over points $\{x_i\}\subseteq X$ -- between $x_i$ and the closest $f(r_m)$, until a suitable convergence criterion is met \cite{pkekalska2006prototype}.

\section{Detection test}
\label{sec:detection-test}

In this paper, we monitor a process generating attributed graphs in order to detect changes in the distribution as well as anomalous events that might occur. 
As we have seen in the previous section, the proposed methodology consists in embedding input graphs into a manifold $\mathbb{M}$, and perform the detection to numeric stream $x_1,x_2,\dots,x_t,\dots$. 
Denote as $F$ the nominal distribution of $x_t$ induced by $Q$ through mapping $f(\cdot)$. Accordingly, the non-nominal distribution of $x_i$ induced by $\widetilde Q$ is denoted by $\widetilde F$.

The detection method requires the evaluation of the mean of random vector $x$ that, for a Riemannian manifold, is defined differently from the Euclidean case. In particular, here, we adopt the Fr\'echet mean $\frmu[F]$ \cite{bhattacharya2017omnibus}:
\begin{equation}
\label{eq:pop-frechet-mean}
\frmu[F] = \argmin_{x\in\mc M} \int_{\mc M} \rho(x,y)^2 dF(y).
\end{equation}
In the case of a finite i.i.d.\ sample $\vec x=\{x_i\}_i$, we can estimate $\frmu[F]$ using its sample counterpart:
\begin{equation}
\label{eq:sample-frechet-mean}
\frmu[\vec x] = \argmin_{x\in\mc M} \sum_{x_i\in\vec x} \rho(x,x_i)^2.
\end{equation}

In order to identify changes, we apply a change detection test inspired by the CUSUM accumulation process. 
At every time step $t$, we compute a statistic $e_t:=\rho(\frmu[F],x_t)$ and sequentially update another statistic $S_t$ that aggregates information from different time steps. Statistic $S_t$ is computed as follows
\begin{equation*}
\label{eq:cusum-iteration}
    \begin{cases}
        S_t = \max_{}\left\{\;0\,,\;S_{t-1}+(e_t-q)\;\right\}\\
        S_0=0,
    \end{cases}
\end{equation*}
where $q$ is a sensitivity parameter. During the nominal regime, statistic $S_t$ is expected to be zero. The hypothesis test is of the form:
$$
\begin{array}{ll}
H_0:\quad \EE[S_t] = 0\\
H_1:\quad \EE[S_t] > 0.
\end{array}
$$
When statistic $S_t$ exceeds a certain threshold $h_t$, a change is detected with a given confidence. The time of the change is 
\begin{equation*}
    \widehat \tau = \inf \{\;t\,:\, S_t>h_t\,\}.
\end{equation*}
Quantity $h_t$ is set to yield a user-defined significance level \begin{equation}
\label{eq:sig-level}
\alpha\;=\;\prob(\,S_t>h_t\;|\;S_i\leq 0, \forall i<t ,\; H_0\,).\end{equation}

The detection of an anomalous graph is performed similarly, but relies on a different hypothesis test. An anomaly is detected at time $\tau$ if the observed $e_\tau$ (same statistic as the one used for the change detection) is larger than a threshold $h$. The critical value $h$ can be estimated according to a user-defined significance level
$$
\alpha= \prob(e>h|e\sim Q).
$$

\section{Experiments}
\label{sec:experiments}

\begin{table*}[t]
\centering
\small
\caption{Comparisons between embedding on constant-curvature manifolds $\mathbb M_\kappa$ w.r.t.\ to a change detection operating directly in the graph domain and on the dissimilarity representation. Tests are performed on different datasets of increasing difficulty (larger numbers indicate more difficult detection problems). Parameters $d$ and $M$ represent the dimension of the manifold and the number of prototypes, respectively. Symbol $\dagger$ highlights when the Euclidean method performed statistically worse than spherical or hyperbolic method. Symbol $\ddagger$ shows when hyperbolic case performed statistically better than spherical one.}
\label{tab:M15p30}
\begin{tabular}{|cccc|rc|cc|cc|cc|}
\hline
\multicolumn{4}{|c|}{Experiment} & \multicolumn{2}{c|}{DCR} & \multicolumn{2}{c|}{$\arl_0$} & \multicolumn{2}{c|}{$\arl_1$}    \\
Embedding & Difficulty   & $M$ & $d$ & mean & 95\% C.I. & mean & 95\% C.I. & mean & 95\% C.I.    \\
\hline
\texttt{Graph Domain} & 2        & - & -  & 1.000  & [1.000, 1.000]        & 124   & [41, 435]     & 1     & [1, 2]        \\
\texttt{Graph Domain} & 4        & - & -  & 0.990  & [0.970, 1.000]        & 124   & [41, 435]     & 6     & [2, 38]       \\
\texttt{Graph Domain} & 6        & - & -  & 0.890  & [0.830, 0.950]        & 124   & [41, 435]     & 33    & [4, 178]      \\
\texttt{Graph Domain} & 8        & - & -  & 0.650  & [0.560, 0.740]        & 124   & [41, 435]     & 85    & [19, 248]     \\
\texttt{Graph Domain} & 10       & - & -  & 0.600  & [0.500, 0.690]        & 124   & [41, 435]     & 114   & [25, 489]     \\
\texttt{Graph Domain} & 12       & - & -  & 0.560  & [0.460, 0.660]        & 124   & [41, 435]     & 115   & [33, 398]     \\
\hline

\texttt{Spherical Man.} & 2       & 30 & 15  & 0.970  & [0.930, 1.000]        & 93    & [37, 239]     & 3     & [1, 10]       \\
\texttt{Spherical Man.} & 4       & 30 & 15  & 0.960  & [0.920, 0.990]        & 93    & [37, 239]     & 6     & [1, 31]       \\
\texttt{Spherical Man.} & 6       & 30 & 15  & 0.970  & [0.930, 1.000]        & 93    & [37, 239]     & 11    & [3, 44]       \\
\texttt{Spherical Man.} & 8       & 30 & 15  & 0.820  & [0.740, 0.890]        & 93    & [37, 239]     & 67    & [14, 317]     \\
\texttt{Spherical Man.} & 10      & 30 & 15  & 0.600  & [0.500, 0.690]        & 93    & [37, 239]     & 90    & [22, 257]     \\
\texttt{Spherical Man.} & 12      & 30 & 15  & 0.580  & [0.480, 0.680]        & 93    & [37, 239]     & 89    & [27, 237]     \\
\hline

\texttt{Euclidean Man.} & 2       & 30 & 15  &$\dagger$ 0.040  & [0.010, 0.080]        & 94    & [40, 209]     & 137   & [5, 369]      \\
\texttt{Euclidean Man.} & 4       & 30 & 15  & 0.980  & [0.950, 1.000]        & 94    & [40, 209]     & 19    & [3, 87]       \\
\texttt{Euclidean Man.} & 6       & 30 & 15  & 1.000  & [1.000, 1.000]        & 94    & [40, 209]     & 21    & [7, 60]       \\
\texttt{Euclidean Man.} & 8       & 30 & 15  &$\dagger$ 0.550  & [0.450, 0.650]        & 94    & [40, 209]     & 86    & [28, 215]     \\
\texttt{Euclidean Man.} & 10      & 30 & 15  &$\dagger$ 0.500  & [0.400, 0.600]        & 94    & [40, 209]     & 107   & [32, 328]     \\
\texttt{Euclidean Man.} & 12      & 30 & 15  & 0.600  & [0.500, 0.690]        & 94    & [40, 209]     & 102   & [31, 312]     \\
\hline

\texttt{Hyperbolic Man.} & 2       & 30 & 15  & 0.960  & [0.920, 0.990]        & 102   & [38, 254]     & 8     & [2, 42]       \\
\texttt{Hyperbolic Man.} & 4       & 30 & 15  &$\ddagger$ 1.000  & [1.000, 1.000]        & 102   & [38, 254]     & 4     & [2, 11]       \\
\texttt{Hyperbolic Man.} & 6       & 30 & 15  & 0.990  & [0.970, 1.000]        & 102   & [38, 254]     & 11    & [4, 48]       \\
\texttt{Hyperbolic Man.} & 8       & 30 & 15  & 0.740  & [0.650, 0.820]        & 102   & [38, 254]     & 73    & [17, 285]     \\
\texttt{Hyperbolic Man.} & 10      & 30 & 15  & 0.690  & [0.600, 0.780]        & 102   & [38, 254]     & 82    & [25, 227]     \\
\texttt{Hyperbolic Man.} & 12      & 30 & 15  & 0.590  & [0.490, 0.690]        & 102   & [38, 254]     & 90    & [28, 276]     \\
\hline
 
 \texttt{Dissimilarity Repr.}  & 2       & 15 & -1  & 1.000  & [1.000, 1.000]        & 68    & [35, 131]     & 1     & [1, 1]        \\
 \texttt{Dissimilarity Repr.}  & 4       & 15 & -1  & 1.000  & [1.000, 1.000]        & 68    & [35, 131]     & 5     & [3, 7]        \\
 \texttt{Dissimilarity Repr.}  & 6       & 15 & -1  & 0.450  & [0.350, 0.550]        & 68    & [35, 131]     & 108   & [16, 441]     \\
 \texttt{Dissimilarity Repr.}  & 8       & 15 & -1  & 0.050  & [0.010, 0.100]        & 68    & [35, 131]     & 262   & [0, 575]      \\
 \texttt{Dissimilarity Repr.}  & 10      & 15 & -1  & 0.040  & [0.010, 0.080]        & 68    & [35, 131]     & 132   & [0, 345]      \\
 \texttt{Dissimilarity Repr.}  & 12      & 15 & -1  & 0.050  & [0.010, 0.100]        & 68    & [35, 131]     & 199   & [0, 560]      \\
\hline
 \end{tabular}
\end{table*}
\begin{figure*}[ht]
\centering
\includegraphics[width=\textwidth]{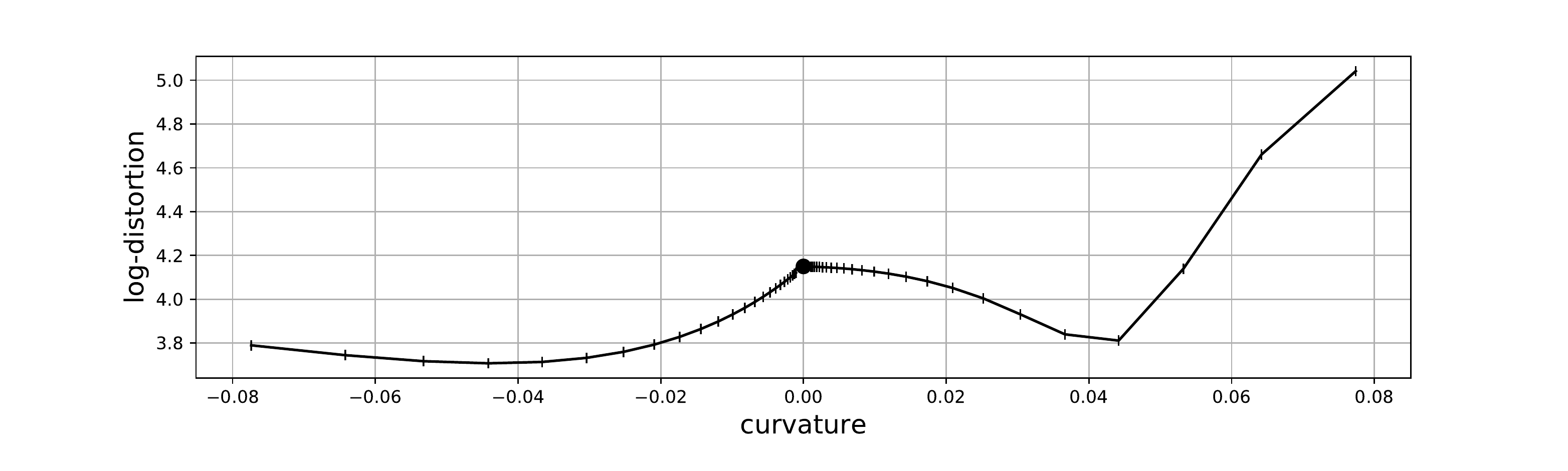}
\caption{Embedding distortion of Equation \eqref{eq:embedding-distortion} observed for different curvatures $\kappa$. The logarithm of the distortion is computed in the first step of the training stage. The curve is composed of three distinct parts, corresponding to the hyperbolic, Euclidean and spherical manifolds. The curve is not continuous in $\kappa=0$, and the central filled circle represents the distortion measured for the Euclidean embedding ($\kappa=0$).
The range of considered curvatures has been selected according to the fact that the maximal distance observable on a $d$-sphere is $\pi\,r$.}
\label{fig:distortion}
\end{figure*}

The experimental campaign is designed to assess the impact of using constant-curvature manifolds as embedding spaces when dealing with anomaly and change detection problems in graph streams.
In this section, the experimentation limits to the change detection problem, being the anomaly detection a simpler one.

In principle, we should select the mapping by estimating curvature $\kappa$ for the problem at hand. However, for comparison and discussion, here we report the performance for three values of $\kappa$: $\kappa=0$, and $\kappa$ estimated on the spherical and hyperbolic manifolds.
These settings are denoted in Table~\ref{tab:M15p30} as \texttt{Euclidean Man.}, \texttt{Spherical Man.}, and \texttt{Hyperbolic Man.}, respectively.

In order to compare results, we applied the change detection test in the graph domain (\texttt{Graph Domain}) as explained later, which we consider as a sort of ``ground truth''. In fact, all embedding methods considered here introduce a distortion of the original pair-wise distances, which we expect to affect change detection results.
We consider also the dissimilarity representation, \texttt{Dissimilarity Repr.}, as further element of comparison \cite{2017arXiv170606941Z}.

Experiments in the graph domain are performed by monitoring statistic $\graphdist(\frmu[\vec g_0], g_t)$, where $\vec g_0$ is a set of training graphs on which we estimated $\frmu[Q]$ and $\graphdist(\cdot,\cdot)$ is the graph edit distance \cite{riesen2013novel}.
The experiments denoted with \texttt{Spherical Man.}, \texttt{Euclidean Man.}, and \texttt{Hyperbolic Man.} consider the numeric $x$-stream where each point $x_t=f(g_t)\in\mc M_\kappa$ is the image of graph $g_t$ embedded through $f(\cdot)$. Here, the monitored statistic is the associated geodesic distance $\rho_\kappa(\frmu[\vec x_0], x_t)$, and $\frmu[\vec x_0]$ is the Fr\'echet sample mean computed w.r.t.\ the manifold geometry, and $\vec x_0 =f(\vec g_0)$.
Finally, \texttt{Dissimilarity Repr.} monitors statistic $\norm{\smean{\vec y}_0-y_t}{2}$, where $y_t$ is the dissimilarity representation of $g_t$ and $\smean{\vec y}_0=\frac{1}{|\vec y_0|}\sum_{y\in\vec y_0} y$ is the ordinary sample mean computed on the dissimilarity representations of the training set.

We remind that, here, we set this particular change detection strategy based on a scalar score only for comparison reasons. In general, attaining a multivariate monitoring would result in a more effective detection. However, implementing such a multivariate test in graph domain is a computationally expensive task as a consequence of the fact we need to implement a graph matching.

The graph streams processed here are based on the Delaunay graph dataset already adopted in \cite{zambon2017detecting}. The dataset contains graphs whose node attributes represent points in the plane, and edges are determined by the Delaunay triangulation of the points. The dataset is composed of several classes. Each class is characterised by an increasing difficulty in distinguishing it from a reference class. Further details can be found in \cite{zambon2017detecting} and are not reported here for the sake of brevity.

\subsection{Figures of merit}

In monitoring time-series for change detection, a common index for performance assessment is the \emph{average run length} ($\arl$). This quantity measures the average number of time steps between consecutive alarms raised by the change detector. One of the advantages is its independence from the length of the input stream.
The average run length assessed during the nominal regime is denoted by $\arl_0$, and can be thought as the inverse of the false alarm rate. Conversely, $\arl_1$ represents the $\arl$ during the non-nominal regime. 

We also considered a further statistic, the detected change rate (DCR) $\in[0,1]$. In order to infer whether or not a change has occurred in a sequence, we estimate both $\arl_0$ and $\arl_1$, then we say that a change is detected if $\arl_0>\arl_1$. 

As DCR approaches 1, the performance of the detector improves, whereas low values of $\arl_1$ depict a prompt detection.
$\arl_0$ is set to $\frac{1}{1-\alpha}$ to yield a user-defined significance level $\alpha$, as in \eqref{eq:sig-level}. Therefore, $\arl_0$ is expected to be constant.

\subsection{Parameters setting}
\label{sec:exp:parameters}

Independently from the particular embedding, we have to compute distances between graphs $\{g_t\}$ and the prototypes $\{r_m\}$. 
Here we used a Graph Edit Distance (GED) algorithm with polynomial complexity based on the Volgenant and Jonker algorithm \cite{riesen2013novel}.
Distances on the Euclidean, spherical, and hyperbolic manifolds are computed by means of equations \eqref{eq:euclidean-distance}, \eqref{eq:spherical-distance}, and \eqref{eq:hyperbolic-distance}, respectively.

We generated one hundred sequences of graphs. Each sequence is produced by bootstrapping graphs from a class $0$ until change time $\tau$ is reached. Then, graphs are bootstrapped from a different class, allowing us to simulate an i.i.d.\ sequence of graphs. 
The first $300$ graphs are reserved to learn the curvature -- i.e., the manifold -- and for prototypes selection. 
Further, $N=600$ graphs are used to train the change detector. 
On these data, the Fr\'echet mean $\frmu[F]$ and the threshold $h_t$ are estimated. In general, we are not able to compute the population mean \eqref{eq:pop-frechet-mean} in the graph domain or on a manifold; hence we estimate it with the sample mean \eqref{eq:sample-frechet-mean}. 
Threshold $h_t$ is set so as to yield a 99\% confidence level for the change detection test. Parameter $q$ is set to the estimated third quartile of $e_t$.

Finally, $2 N$ graphs have been generated to emulate the operational phase.
In particular, the first $N$ graphs are produced according to the nominal distribution, whereas the second $N$ simulate the change and hence are representative of the non-nominal regime.

\subsection{Results}

Table~\ref{tab:M15p30} shows the results of the experiments. 
First of all, the 95\% confidence intervals related to the estimated $\arl_0$ provide evidence that threshold learning completed successfully; indeed, the 95\% confidence intervals contain the target value of $\arl_0$, here $100$. Notice that the $\arl_0$ estimations are identical when computed within the same embedding. This is because a seed for the pseudo-random generator has been set for experiment reproducibility.

In Table~\ref{tab:M15p30}, we observe that different curvatures for the embedding space produce different DCR's. Our results show that the {hyperbolic and spherical} spaces performed better than the {Euclidean} one in the considered application.
On the other hand, spherical and {hyperbolic} embeddings attain comparable results and, in one case, the {hyperbolic} embedding outperformed the {spherical} one.

In Figure~\ref{fig:distortion}, the distortion \eqref{eq:embedding-distortion} introduced by the embedding is analysed as a function of the curvature. 
{In the region of positive curvature, we spot a local minimum around $\kappa=0.04$. Looking at negative curvatures, there is a minimum around $\kappa=-0.05$ which is lower than the distortion at $\kappa=0.04$. 
The Euclidean case, with $\kappa=0$, produces a distortion worse than the hyperbolic and spherical ones. The behaviour observed in terms of distortion \eqref{eq:embedding-distortion} is concordant with the detection rates in Table~\ref{tab:M15p30}; notice, however, that the distortion assesses the goodness of the embedding limited to the training set, and hence is not directly related to the change detection performance.
We observe also that, despite the three types of embedding are obtained by different optimisation problems \eqref{eq:common-opt-problem} and \eqref{eq:opt-problem-b}, the distortion appears to be continuous around $\kappa=0$.

Finally, we conclude that the nominal class turned out to favour embedding on curved spaces, in particular those with negative curvatures. However, the results reported here focus only on a particular dataset and hence more experiments are needed to confirm these findings. 
}

The method \texttt{Graph Domain} appears, in general, {to perform well} even though is not all differences are statistically significant. 
Regarding \texttt{Dissimilarity Repr.}, the detection is effective as far as the problem is sufficiently simple. Once the two distribution $F$ and $\widetilde F$ have the same support, the detection fails completely. Results are different from the ones published in \cite{zambon2017detecting}; this is justified by the different experimental setting adopted here.
First of all, the change detection test in \cite{zambon2017detecting} is a multivariate one, which increases the amount of information extracted and monitored from the graph stream. As described above, the scalar setting adopted in the present paper is necessary to attain a fair comparison with \texttt{Graph Domain}. 
Secondly, the multivariate change detection test monitors windows of data making possible the exploitation of a known distribution and, consequently, setting virtually exact thresholds. 

Overall, we conclude that when the mean graphs of the two distributions, $Q$ and $\widetilde Q$, are far apart, then simpler approaches, like \texttt{Dissimilarity Rep.}, are more effective than manifold-based ones. In fact, the manifold is learned on the nominal distribution, hence it may poorly generalise to very diverse graphs. Conversely, when the problem at hand gets more challenging and the two distributions almost overlap (as in smooth drift type of changes), we observe the effectiveness in change detection problems of embedding onto manifolds of constant, non-zero curvature.

\section{Conclusions}
\label{sec:conclusions}

Performing change detection in graph domains is a challenging problem from both a theoretical and computational point of view.
Recently, we have proposed a methodology to perform change detection on sequences of graphs based on an embedding procedure \cite{2017arXiv170606941Z}: graphs are mapped to numeric vectors so that change detection can be performed on a standard geometry setting. The embedding was realized by means of the dissimilarity representation.
In this paper, we elaborated over our previous contribution and, in particular, we performed embedding onto three constant-curvature manifolds having planar, spherical, and hyperbolic geometry. Our motivation comes from the fact that complex data representations such as graphs might not be described by a simple Euclidean structure. This intuition is corroborated by several recent results (e.g., see \cite{wu2015emergent,boguna2010sustaining}) suggesting that the geometry underlying complex networks can have a hyperbolic geometry.

Our results indicate that, in the first place, (i) varying the manifold curvature can reduce the distortion of the distances; furthermore, the curvature can be treated as learning parameter for adapting the manifold to the specific application of interest in a data-driven fashion. In the second place, (ii) embedding on such manifolds can be effectively employed to detect change events and anomalies in a stream of graphs. In particular, results showed that the performance of the proposed change detection method is comparable to the corresponding ground-truth version operating directly on graphs.

Future directions include a better experimental evaluation of the proposed embedding procedure on constant-curvature manifolds, accounting also for analytical solutions to the involved optimisation problems and automatic optimisation of relevant parameters. In addition, we plan to work on theoretical aspects related to asymptotic distribution estimations on manifolds and their application in change detection for graphs.

\section*{Acknowledgements}

This research is funded by the Swiss National Science Foundation project 200021\_172671: ``ALPSFORT: A Learning graPh-baSed framework FOr cybeR-physical sysTems''.

\bibliographystyle{ieeetr}
\bibliography{sample}

\end{document}